\documentclass{article}

\usepackage{arxiv}

\usepackage[utf8]{inputenc} 
\usepackage[T1]{fontenc}    
\usepackage{hyperref}       
\usepackage{url}            
\usepackage{booktabs}       
\usepackage{amsfonts}       
\usepackage{nicefrac}       
\usepackage{microtype}      
\usepackage{fancyhdr}       
\usepackage{graphicx}       
\graphicspath{{media/}}     
\usepackage{amsmath}
\usepackage{amsthm}
\usepackage{bbm, dsfont}
\usepackage{algorithm2e}
\usepackage{subcaption}
\usepackage{cleveref}
\usepackage{xcolor}
\usepackage{hyperref}
\usepackage{dblfloatfix}
\newtheorem{thm}{Theorem}
\newtheorem{definition}{Definition}

\title{Ex-Ante Assessment of Discrimination in Dataset

}

\author{
  Jonathan Vasquez\\
  George Mason University \\
  Universidad de Valparaiso \\
  \texttt{jvasqu6}@gmu.edu \\
  \texttt{jonathan.vasquez}@uv.cl \\
   \And
   Xavier Gitiaux \\
   George Mason University \\
    \texttt{xgitiaux@gmu.edu} \\
   \And
   Huzefa Rangwala \\
    George Mason University \\
   \texttt{rangwala}@gmu.edu \\
}

\begin{document}
\maketitle

\begin{abstract}
Data owners face increasing liability for how 
the use of their data could harm under-priviliged communities.  Stakeholders would like to identify the characteristics of data that lead to algorithms being biased against any particular demographic groups, for example, defined by their race, gender, age, and/or religion. Specifically, we are interested in identifying subsets of the feature space where the ground truth response function from features to observed outcomes differs across demographic groups. To this end, we propose \emph{FORESEE}, a \emph{FORES}t of decision tr\emph{EE}s algorithm, which generates a score that captures how likely an individual’s response varies with sensitive attributes. Empirically, we find that our approach allows us to identify the individuals who are most likely to be misclassified by several classifiers, including Random Forest, Logistic Regression, Support Vector Machine, and k-Nearest Neighbors. The advantage of our approach is that it allows stakeholders to characterize risky samples that may contribute to discrimination, as well as, use the \emph{FORESEE} to estimate the risk of upcoming samples.
\end{abstract}

\keywords{algorithmic fairness \and discrimination risks \and bias identification \and ex-ante assessment}

\section{Introduction}
\label{sec:introduction}
A growing concern with algorithmic decision making systems is that they might replicate or exacerbate existing social biases. Empirical evidence show that algorithmic outcomes might depend on demographic characteristics (e.g. gender or race) that should be irrelevant to the task. Examples span many domains, including criminal justice \cite{green2019disparate, ProPublica2016}, banking and finance \cite{fu2021crowds, wen2019fairness}, health-care \cite{chen2021algorithm, seyyed2020chexclusion}, education \cite{kizilcec2020algorithmic} and facial recognition \cite{buolamwini2018gender}. Scholars and practitioners have developed toolkits aimed at measuring and mitigating unfairness outcomes \cite{aif360-oct-2018, bird2020fairlearn}. 
However, these might not be suitable for early steps of machine learning pipelines, where data-owners without any control of the down-stream development might want to remove any information regarding sensitive attributes \cite{gitiaux2021fair}, or complement the data delivery with context-aware components that warn potential unfairness issues.

In machine learning pipelines \cite{alla2021mlops}, unfairness issues are commonly identified after models are trained, validated, and sometimes deployed \cite{pessach2020algorithmic}. To reduce its negative impacts, developers may use pre-processing, in-processing, and/or post-processing mitigation approaches \cite{akpinar2022sandbox}. 
Moreover, authors have argued that root causes for unfair algorithm outcomes include social or historical biases encoded in the data -- data collections that over or under-represent a minority \cite{chen2018my,ensign2018runaway} -- and heteroskedastic noise between demographics groups \cite{chen2018my}. Consequently, auditing for unfairness in the early steps will lead to effective and efficient mitigation of unfairness outcomes. To this end, Datasheets \cite{gebru2021datasheets}, Method Cards \cite{adkins2022prescriptive}, and Model Cards \cite{mitchell2019model} help provide warnings for potential misuse in downstream algorithms and decision making. This paper contributes to this research stream by 
proposing an ex-ante evaluation of unfair outcomes risk that enriches the warnings flags for down-stream developments.

Our notion of ex-ante risk measures how the ground truth mapping from features to label varies with the demographic characteristics of each individual. We argue that our measure of risk captures systematic differences in the data generating process between demographic groups and that these differences are 
predictors of whether future uses of the data in a machine leaning pipeline will lead to outcomes that violate standard measures of fairness. Moreover, we show that we can obtain reasonable estimates of our ex-ante measure of risk by either boosting or bagging decision trees. Boosting is based on Bayesian Additive Regression Trees (BART, \cite{chipman2010bart}). Bagging relies on our own variant of BART (named as FORESEE, a FORESt of decision trEEs algorithm), where we ensemble decisions trees trained on different combinations of data features. Our intuition is that local estimation is preferable for unfair risk evaluation instead of using response functions learned over dataset. Using numerical simulations, we show that FORESEE outperforms BART and generates statistically unbiased estimates for our ex-ante risk. 

We apply our measure of ex-ante risk to three benchmarks datasets. (1) We find that our ex-ante measure of risk does correlate with standard measurement of unfair outcomes, including demographic parity, equalized opportunity, and equalized odd. 
(2) Sub-populations with high ex-ante risk are more likely to be subjected to high unfairness by classifiers trained on a sample of the data. Therefore, our ex-ante risk measure provides useful warning to data owners and stakeholders before the data is ingested by a classifier. In a context where organizations that own data are increasingly liable for future unfair uses of the data, our ex-ante risk could be instrumental to decide whether a data can be distributed to machine learning pipelines. Moreover, it allows identifying the characteristics of samples that are the most likely to be exposed or contribute to unfair outcomes and thus, guiding future data collection and model development.   

Our contributions are as follows:
\begin{itemize}
\item We formalize stress testing  data for potentially discriminatory future uses as the problem of measuring differences in data generation processes across demographic groups.
\item We provide experimental proofs of how our measure of ex-ante risk is a useful warning sign for the demographic disparity, inequality of odds and opportunities of classifiers trained on a sample of the data.
\item We show how model averaging over diverse decision trees provides reliable estimates for our ex-ante risk. 
\end{itemize}

\section{Problem Setting}
\label{sec:problem_setting}
We frame the task where a set of individuals sampled from $\mathcal{X}\times\mathcal{Y}$ (where $x\in \mathcal{X}$ describes individual's features and $y\in \mathcal{Y}$ the ground truth of target variable) can lead into finding $g: \mathcal{X}\mapsto\mathcal{Y}$ used to support decision-making process. We assume that $g$ has a sufficient performance such that the decision-makers consider the function $g$ as part of the process. For example, $g$ can be the classifier for mortgage applications, where upcoming applicants' information are the inputs and low and high risk of mortgage delinquency is the outcome. Additionally, we assume that for each sampled individual we are given their sensitive attributes $s \in \mathcal{S}$, which represents protected information forbidden to be used in the classification. Furthermore, sensitive attribute is defined by the union of the different groups, i.e., $S = \cup_j S_i \forall i\in[j]$, with $j$ the total of different groups and $S_i$ is a group of individuals with same sensitive attribute(s) usually represented by a number. Although $S$ is not part of the inputs, they are allowed to be used as the evaluation of discrimination across the groups in the sensitive attributes.

The objective is to determine how likely any individual $(x,s,y) \in \mathcal{D}$ will lead into discrimination a function $g:\mathcal{X}\mapsto\mathcal{Y}$ learned from $\mathcal{D}$ s.t. $g$ show sufficient performance to be used by decision-makers. This objective implies two challenges. First, we have to define what \emph{discrimination} stands for. And second, we are aimed to determine the discrimination when $g$ is unknown, i.e., under ex-ante notion. We discuss and address these two aspects in the following section under a binary classification task, although this is extensible to other task types.

\subsection{Risk Scores and Aggregate Misclassification Rates}
Discrimination can be understood as the disparity between two groups of individuals. In the mortgage application example, disparities can be found along the process, i.e., if there is explicit evidence of disparity through the use of sensitive attributes as part of the classification, and there is sign of dependency between the outcome and the protected attribute, it can be determined that there exists disparate treatment \cite{feldman2015certifying}. One solution is to remove any sensitive attribute, which is known as Fairness Through Unawareness (FTU) \cite{dwork2011fairness}. However, FTU is not enough since the correlation of sensitive attribute with other features may still end up with disparities in the outcome, which is known as disparate impact \cite{feldman2015certifying}. Based on this fairness notion, Demographic Parity ($demP$) measures disparity across groups over the outcome of a function $g$ \cite{feldman2015certifying}. Similarly, individual fairness certificates that $g$ maps similar outcomes for similar individuals \cite{ilvento2019metric}. Using these notions, we define an \emph{ex-ante} measure by evaluating disparity on the labeled dataset (instead of the outcome of $g$) for similar individuals of a given sample. More in detail, we state that \emph{fairness risk score} is the conditional probability of $Y$ being $1$ (from $demP$) given $X=x$ with differences in the sensitive attribute $S$ (from individual fairness).

In the following formulation, we first discuss theoretically how the risk can be obtained from infinite population (or at least large enough). However, since this is not true for most of the cases, we then discuss two estimation proposals under finite population in Section \ref{subsec:estimation_risk}.

\begin{definition}
\label{def:risk_score}
For any individual $(x, s)\in \mathcal{X}\times\mathcal{S}$, its fairness risk score $r(x)$ is defined as
\begin{equation}
    r(x) = |P(Y=1|S=s, X=x) - P(Y=1|S \neq s, X=x).
\end{equation}
\end{definition}

We claim that \textbf{Definition \ref{def:risk_score}} is ex-ante since it does not depend on any classifier that will use the data to make decisions. However, this is based only on \textit{independence} notions of disparate treatments, which can lead into flaws of identification of discrimination. According to Kizilcec and Lee \cite{kizilcec2020algorithmic}, \emph{separate} notion can solve this issue by including the ground truth. We argue that \textbf{Definition 1}, as ex-ante risk, captures this granular and aggregate fairness properties of classifiers using the data: (i) differences in individual misclassification rates across demographic groups; (ii) differences in aggregate misclassification rates across demographic groups. Moreover, we show that ex-ante risk score is directly related to the differences $\delta_{mis}(x)$ in misclassification rate conditional on $X=x$ for an unknown function $g$, where $\delta_{mis}$ is defined as Equation \ref{delta:mis}.

\begin{equation}
\label{delta:mis}
    \delta_{mis}(x) = |P(g(X)\neq Y | X=x, S=s) - P(g(X)\neq Y | X=x, S\neq s)|.
\end{equation}

Having this we state the Theorem \ref{thm: 1}, which implies that the ex-ante risk score captures how the misclassification rate of any unaware classifier vary across demographic groups. Consequently, for a given $X=x$, a large risk score means that individuals in one group will be more likely to be misclassified than in the other, regardless of the classifier.

\begin{thm}
\label{thm: 1}
For any unaware classifier $g$, $x\in \mathcal{X}$, we have:
\begin{equation}\nonumber
    \delta_{mis}(x) = r(x)\left|I_{\{g(x)=0\}} - I_{\{g(x)=1\}}\right|, 
\end{equation}
 where $I_{\{g(x)=1\}}, I_{\{g(x)=0\}}$ are the characteristic functions of the sets $\{x\in \mathcal{X}|g(x)=1\}$ and $\{x\in \mathcal{X}|g(x)=0\}$. Moreover, for a deterministic unaware classifier, $\delta_{mis}(x)=r(x)$.
\end{thm}

\textit{Proof}. Let $\eta_{s}(x)$ denote $P(Y|X=x, S=s)$. For any classifier $g$, we have

\begin{equation*}
    \begin{split}
        P(g(X) = Y| X=x, S=s) & =  P(Y=1, g(X)=1|X=x, S=s) + P(Y=0, g(X)=0|X=x, S=s) \\
        & \overset{(a)}{=} I_{\{g(x)=1\}}\eta_{s}(x) + I_{\{g(x)=0\}}(1-\eta_{s}(x)),
    \end{split}
\end{equation*}

where $I_{\{g(x)=1\}}$ is the indicator function equal to one whenever $g(x)=1$; and, (a) uses the fact that conditional on $X$, $g(X)$ is independent of $Y$. Therefore, 
\begin{equation*}
\begin{split}
    \delta_{mis}(x)  & = |I_{\{g(x)=1\}}(\eta_{0}(x)-\eta_{1 }(x)) - I_{\{g(x)=0\}}(\eta_{0}(x -\eta_{1}(x))| \\
    & = r(x)\left|I_{\{g(x)=1\}} - I_{\{g(x)=0\}}\right|.
    \end{split}
\end{equation*}

If $g$ is deterministic, we have either $g(x)=0$ or $g(x)=1$, which leads to $\delta_{mis}(x)=r(x)$.

Our ex-ante fairness risk score also captures aggregate fairness properties of any classifier $g$, which allows us to look at the aggregate difference in misclasification rates \cite{chen2018my} defined as Equation \ref{delta:mis_agg}.

\begin{equation}
\label{delta:mis_agg}
    \Delta_{mis}(g) = |P(g(X)\neq Y|S =s) - P(g(X)\neq Y|S\neq s)|.
\end{equation}

Large values for $\Delta_{mis}$ means that the error rate is not equalized across groups. Having this, we state Theorem \ref{thm: 2}, which implies that if the distribution of features $X$ does not depend on the sensitive attribute, then the total variation between distribution of $X$ conditional on $S=s$ and $S\neq s$ is zero and low aggregate risk scores implies low $\Delta_{mis}(g)$. Therefore, whenever the distribution shift between $X|S=s$ and $X|S\neq s$ is small, our measure of risk allows users to identify data that are unlikely to lead to unfair classification, when fairness is measured in terms of difference in misclassification rates (i.e., in terms of \emph{separation} notions \cite{kizilcec2020algorithmic}).
 
\begin{thm}
\label{thm: 2}
For any unaware classifier $g$, 
\begin{equation}
    \Delta_{mis}(g) \leq
    \frac{1}{2}E_{x\sim P_{s}}[r(x)] + \frac{1}{2}E_{x\sim P_{ns}}[r(x)] + TV(P_{s}, P_{ns}),
\end{equation}
where $TV(P_{s}, P_{ns})$ is the total variation between the distribution $P_{s}$ of $X$ conditional on $S=s$ and the distribution $P_{ns}$ of $X$ conditional on $S\neq s$.
\end{thm}

\emph{Proof}. Let $\eta_{s}(x)$ denote $P(Y|X=x, S=s)$:
\begin{equation*}
    \begin{split}
      P(g(X)= Y | S=s) & = \displaystyle\int P(g(X)= Y| S=s, X=x) P_{s}(dx)  \\
      & = \displaystyle\int \left[I_{\{g(x)=1\}}\eta_{s}(x) \right. + \left. I_{\{g(x)=0\}}(1 -\eta_{s}(x)) \right] P_{s}(dx).
    \end{split}
\end{equation*}

Let $c(x)$ denote $I_{\{g(x)=1\}} - I_{\{g(x)=0\}}$. Therefore, we have

\begin{equation*}
    \begin{split}
        \Delta_{miss}  = & \left|\displaystyle\int c(x)\eta_{s}(x)P_{s}(dx) - \displaystyle\int c(x)\eta_{ns}(x)P_{ns}(dx)  \right| \\
         = &\left|\displaystyle\int c(x)\frac{\eta_{s}(x) + \eta_{ns}(x)}{2} \left[P_{s}(dx) - P_{ns}(dx)\right]\right. + \displaystyle\int c(x)\frac{\eta_{s}(x) - \eta_{ns}(x)}{2} P_{s}(dx) \\
        &  + \left.\displaystyle\int c(x)\frac{\eta_{s}(x) - \eta_{ns}(x)}{2} P_{ns}(dx) \right| \\
         \overset{(a)}{\leq} & \displaystyle\int \frac{r(x)}{2} P_{s}(dx) + \displaystyle\int \frac{r(x)}{2} P_{ns}(dx) + \displaystyle\int\left|P_{s}(dx) - P_{ns}(dx)\right|,
    \end{split}
\end{equation*}

 where $(a)$ uses the fact that $c(x)\leq 1$ and $\eta_{s}, \eta_{ns}\leq 1$, the triangular inequality. The result in Theorem \ref{thm: 2} follows from the definition of total variation.

\subsection{Risk Scores and Sub-population Fairness Metrics}
Results in Theorem \ref{thm: 2} do not provide us with a lower bound on aggregate fairness metrics, even in the absence of a distributional shift between $P_{s}$ and $P_{ns}$. In this section, we show a lower bound for misclassification rates at the sub-population level. Specifically, for $\gamma\in(0, 1)$, a sub-population level fairness metric measures the maximum value this metric takes over any subset $G$ of the feature spaces $\mathcal{X}$ such that $P(G)\geq \gamma$ (e.g. \cite{gitiaux2019multi,kearns2018preventing,kim2019multiaccuracy}).  Sub-population level definitions of algorithmic fairness are stronger than their aggregate counterparts, since they protect subgroups defined by a complex intersection of many sensitive attributes (\cite{kearns2018preventing,kearns2019empirical}) or a structured slicing of the feature space \cite{gitiaux2019multi}. We define sub-population differences in misclassification rates of a classifier $g$ as 
\begin{equation}
  \Delta_{sub-mis}(g, \gamma) = \max_{G: P(G)\geq \gamma}\Delta_{mis}(g|x\in G),
\end{equation}
where $\Delta_{mis}(g|x\in G)$ is the difference in misclassification rates between demographic groups in sub-population G. 

\begin{thm}
\label{thm: 3} Suppose that there exists a sub-population $G\subset \mathcal{X}$ such that $P(G)> \gamma$, $P_{s}=P_{ns}$, and $P(Y=1|X=x, S=s) > P(Y=1|X=x, S\neq s)$. For any stochastic classifier $g: \mathcal{X}\rightarrow [0,1]$ such that $inf_{x\in G} g(x)> 1/2$, there exists $\kappa>0$ such that
\begin{equation}
    \Delta_{sub-mis}(g, \gamma) > \kappa E[r(x)|x\in G]. 
\end{equation}
\end{thm}

Theorem \ref{thm: 3} implies that whenever the response function $E[Y=1|X=x, S=s]$ is larger than $E[Y=1|X=x, S\neq s]$ for a sub-population $G$, a classifier that predicts $Y=1$ on this sub-population will lead to higher misclassification rates for one demographic group and the differences in misclassification rates will be bounded from below by the average risk score. Therefore, by looking at sub-populations with high predicted risk scores, the user can identify sub-populations where differences in sensitive attributes $S$ will correlate with differences in misclassification rates.

\emph{Proof}. Let $2\kappa=inf_{x\in G}g(x) - 1/2 > 0$. Using derivations from the proof of Theorem \ref{thm: 1} and the assumption that $P_{0}=P_{1}$, we can show that 
\begin{equation*}
    \begin{split}\Delta_{mis}(g|x\in G) & = \left|\displaystyle\int_{x\in G} (2g(x)-1)(\eta_{s}(x)-\eta_{ns}(x)P(dx) \right| \\
    & \overset{(a)}{>} \kappa \displaystyle\int_{x\in G}\left[\eta_{s}(x) - \eta_{ns}(x) \right]P(dx) \\
    &= \kappa E[r(x)|x\in G],
    \end{split}
\end{equation*}
where $(a)$ uses that $\eta_{s}(x) \geq \eta_{ns}(x)$ on $G$.

\section{Risk Estimation}
\label{subsec:estimation_risk}

To estimate the ex-ante risk score, we propose two methods: (i) a Bayesian-based approach to estimate the probability $P(Y=1|X=x, S=s)$ followed by the computation of the risk score $r(x)$; (ii) an ensemble approach that averages the risk estimates from diverse decision trees.

\subsection{Method 1: Risk Score of Average Models}
Our first approach consists of first obtaining an estimate of the response function $E[Y|X=x, S=s]$ and $E[Y|X=x, S\neq s]$ and then, compute the risk score. We follow \cite{hill2011bayesian} and use an additive tree model BART (\cite{chipman2010bart} to obtain a non parametric estimate of the response functions. Given $M$ trees $T_{1}, T_{2}, ..., T_{M}$, we compute the risk score as 
\begin{equation*}
    r_{BART}(x)=\frac{1}{M}\left|\displaystyle\sum_{m=1}^{M}E(Y|T_{m}, X=x, S=s)\right.-\left.E(Y|T_{m}, X=x, S\neq s)\right|
\end{equation*}

BART uses a prior to regularize the depth of each tree $T_{m}$. This non-parametric approach has been successful to estimate individual treatment effect following the seminal work of \cite{hill2011bayesian}. 

\subsection{Method 2: Average of Risk Scores}
One limitation of a BART estimate of the ex-ante risk score is that it requires to estimate both response functions $E[Y|X=x, S=s]$ and $E[Y|X=x, S\neq s]$, while we are only interested in the difference $E[Y|X=x, S=s] -E[Y|X=x, S\neq s]$. BART averages decision trees to obtain a robust Bayesian estimate of each response function. In this section, we present an algorithm that directly averages the risk score estimates from each decision tree. 

Given a collection of trees $T_{1}, T_{2}, ..., T_{M}$, we define as $L_{1}(x)$, $ L_{2}(x)$, ..., $L_{M}(x)$ the leafs to which an individual with feature $x\in \mathcal{X}$ belongs to. For each decision tree $T_{m}$, we compute its estimate $r(x, T_{m})$ of the risk score as the difference between demographic groups in the estimation error rate $Y_{mis}$  in $L_{m}(x)$:

\begin{equation*}
    r(x, T_{m}) = |E(Y_{mis}|L_{m}(x), S=s) - E(Y_{mis}|L_{m}(x), S\neq s)|
\end{equation*}

Our risk estimate is then 
\begin{equation}\nonumber
    r_{FORESEE}(x) = 
    \frac{1}{M}\displaystyle\sum_{m=1}^{M} r(x, T_{m}).
\end{equation}
Intuitively, decision trees partition the feature space into regions where outcomes are assumed to be constant. Leaf level risk scores characterize how violations of this assumption vary across demographic groups. 

To obtain a robust risk estimate, we average a diverse set of trees obtained as followed: (i) we train each tree on a different random sub-sample of the insances and a random subset of the features; (ii) we retain the trees that achieve a minimum performance $\beta \in (0,1)$. Step (i) allows to ensemble trees that see different aspect of the data generating process and de-correlates estimation errors. Step (ii) allows to retain only the trees that captures at least partially the characteristics of the response function $E[Y|X=x]$. It is worthy to note that we use $S$ in the training. We argue that for a leaf with none members of one group would reflect dependency between $Y$ and $S$ given $X=x$, hence a high risk of discrimination. Under an optimistic scenario, the expected misclassification rate should be equal across groups in the leaf, while the highest error rate is expected for the not present group under pessimistic one. Since we aim to flag probable discrimination, we argue that the latter is preferable. Note that a better approach might be used, however, according to the results in experiments section, our assumption seems to be sufficient. Finally, more details about the \textit{FORESEE} algorithm are provided in the appendix.

\section{Experimental Evaluation}
\label{sec:method}
We design our experiments to answer the four following research questions.

\begin{itemize}
    \item \textbf{RQ1}: Do  risk estimates from \emph{FORESEE} and \emph{BART} average to the ground truth risk in numerical simulations?
    \item \textbf{RQ2}: Is the ex-ante risk score a good predictor of how unfair classifiers using the data as input will be?
    \item \textbf{RQ3}: Does our measure of risk allow the user to describe the sub-populations that are the most likely to be under-served by classifiers using the data as inputs? 
    \item \textbf{RQ4}: How can ex-ante risk score guide the development of fair algorithms? 
\end{itemize}

\subsection{Data sets}
\label{subsec:data_sets}
\paragraph{Synthetic Data}
First, we rely on numerical simulations to test whether FORESEE and BART are on unbiased estimator of the true risk score. We uniformly draw features from $[0,1]^{2}$ and assign a sensitive attribute $s$ with $P(S=1)=0.5$. For $S=1$, we assign a label $Y=1$. For $S=0$, we assign a label $Y=1$ with probability equal to $1 - \frac{x_{1}+x_{2}}{2}$. Therefore, for a given $X=x$, the ground truth risk is equal to $r(x)=\frac{x_{1}+x_{2}}{2}$. We draw $5000$ samples from this synthetic data, run FORESEE and BART to estimates $r_{FORESEE}$ and $r_{BART}$ for each sample and compare their values to the ground truth risk $r(x)$. We repeat the protocol $20$ times with different seeds to compute mean and standard deviation of the risk estimates. 


\paragraph{Real world datasets}
We use two benchmark data sets in fair machine learning and a new dataset in education data mining.  The former two are publicly available at ProPublica webpage and UCI Machine Learning repository. 

\textbf{Adults} (a.k.a. Census Income Data Set) consists of $48,844$ individuals described by $14$ features that include marital status, education, working hours. The sensitive attribute is the gender to which individuals self-identify to. The outcome is whether an individual's income is larger than $50K$ \cite{asuncion2007uci}.

\textbf{Compas} gathers information on $7,214$ individuals in the criminal justice system to assess their recidivism risk. Features include number of felonies, misdemeanors and degree of charges. The sensitive attribute is whether an individual self-identifies as African American. The outcome is whether an individual re-commits a crime within two years \cite{ProPublica2016}.  

\textbf{Dropout} contains academic records of $4,706$ students in a Latin-American university. Features include grades and participation to academic support programs. The sensitive attribute is the gender to which each student self-identifies to. The outcome reflects drop out within the first two years. 

We split each dataset into training and testing sets using a ratio of 70/30. 

\subsection{Experiments Protocol}
\label{subsec:protocol}
\subsubsection{Risk Scores vs. Downstream Fairness Metrics}
To explore how ex-ante risk score is informative of future unfair uses of the data, we train four models -- Logistic Regression (LR), Random Forest (RF), k-Nearest Neighbors (KNN), and Support Vector Machine (SVM) -- to predict $Y$ from $X$. We compute accuracy for evaluating models' performance in Compas, and F-1 score in Adults and Dropout. We also compute three standard fairness metrics: demographic disparity $\delta_{demP}$ \cite{dwork2011fairness}; inequality of opportunity $\delta_{opp}$ \cite{hardt2016equality}; and, inequality of odds $\delta_{odd}$ \cite{hardt2016equality}. We also classify individuals into High and Low Risk whose risk score are larger and smaller than $\lambda$, respectively. 
However, this threshold is truly context dependent. We leave it to the stakeholders to decide which threshold is appropriate in a specific application. 

\subsubsection{Risk Score and Mitigation}
To show how our ex-ante risk score can guide machine learning developers, we explore two risk-based fairness mitigation strategies. We identify all instances with risk score larger than $\lambda$ and apply two mitigation approaches: (i) pre-processing and (ii) post-processing. 

Preprocessing removes the high-risk instances from the training  set and only report classification outcomes on instances with low ex-ante risk score.  This is a very relevant approach whenever stakeholder can replace automated decisions by human ones for outcomes with high ex-ante risk score. On the other hand, post-processing consists of adjusting the classification threshold over the classifier's predicted class conditional probabilities, while minimizing one of the aggregate fairness metrics and the error rates \cite{hardt2016equality}.

\section{Results and Discussions}
\label{sec:results}
\subsection{RQ1: Performance of Risk Score Estimators}
\label{susec:algo_perfo}
\begin{figure}[h]
    \centering
    \begin{subfigure}{0.45\textwidth}
        {\includegraphics[width=\textwidth]{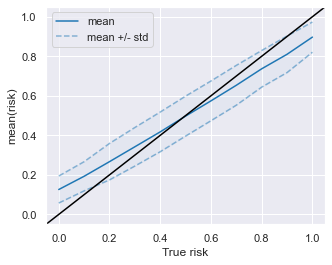}}
        \caption{FORESEE risk estimates}
        \label{fig:MSE_alpha}
    \end{subfigure}
    \begin{subfigure}{0.45\textwidth}
        {\includegraphics[width=\textwidth]{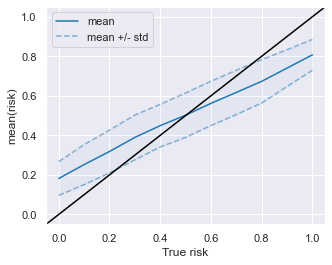}}
        \caption{BART risk estimates}
        \label{fig:MSE_beta}
    \end{subfigure}
    \caption{Means and standard deviations of estimated versus ground truth ex-ante risk scores for FORESEE (left) and BART (right) in our synthetic dataset. This shows that FORESEE is an unbiased estimator of ex-ante risk scores since its mean aligns with the $45^{o}$ line.}
    \label{fig:perfo_synthetic}
\end{figure}

Figure \ref{fig:perfo_synthetic}  shows the mean and standard deviation of  $r_{FORESEE}(x)$ (left) and $r_{BART}(x)$ (right) against the ground truth risk $r(x)$ for each $x$ in our synthetic dataset. 
We observe that  FORESEE generates a statistically less unbiased estimate of the ground truth risk since the mean of the $r_{FORESEE}(x)$ across $20$ simulations is equal to $r(x)$ for almost all values of $r(x)\in(0,1)$.  On the other hand,  BART estimates have an upward bias for low values of the true risk score $r(x)$ and a downward bias for large values $r(x)$. This observation confirms our intuition that it is preferable to directly average risk estimates from local approximations instead of first estimating the response functions over the entire dataset and then computing the risk score.

\subsection{RQ2: Risk Scores vs. Downstream Fairness Metrics}
Table \ref{table:ml_performances} shows the classification performance of four models across our three real world datasets. Best classifiers in terms of F1 score/accuracy vary across datasets: Random Forest achieves the highest performance on Adults and Compas, while Logistic Regression on Dropout. Moreover, best classifiers generate significant fairness metrics, where of particular interest is whether our ex-ante risk score  anticipates these fairness issues.

\begin{table}[h]
\centering
\begin{tabular}{@{}lllll@{}}
\toprule
  \textbf{Model}  & \textbf{Adult} & \textbf{Dropout} & \textbf{Compas}  \\ 
  	& F-1 \textbf{ }$|$ $\delta_{opp}$ $|$ $\delta_{odd}$ $|$ $\delta_{demP}$ &  F-1 \textbf{ } $|$ $\delta_{opp}$ $|$ $\delta_{odd}$ $|$ $\delta_{demP}$ & acc \textbf{ } $|$ $\delta_{opp}$ $|$ $\delta_{odd}$ $|$ $\delta_{demP}$ \\ \midrule
  LR  	& .618\textbf{} $|$ .409\textbf{} $|$ .321\textbf{} $|$ .351	       & \textbf{.655} $|$ .062\textbf{} $|$ .038\textbf{} $|$ \textbf{.031}	& .655\textbf{} $|$ .221\textbf{} $|$ .189\textbf{} $|$ .213	\\
  RF    & \textbf{.703} $|$ \textbf{.200} $|$ .208\textbf{} $|$ .331	       & .614\textbf{} $|$ .069\textbf{} $|$ .045\textbf{} $|$ .048 	        & \textbf{.668} $|$ .178 $|$ .174\textbf{} $|$ .204	\\
  KNN 	& .644\textbf{} $|$ .208\textbf{} $|$ \textbf{.148} $|$ \textbf{.199}  & .551\textbf{} $|$ .085\textbf{} $|$ .050\textbf{} $|$ .041 	        & .650\textbf{} $|$ \textbf{.177} $|$ \textbf{.167} $|$ \textbf{.193}	\\
  SVM  	& .666\textbf{} $|$ .212\textbf{} $|$ .247\textbf{} $|$ .382	       & .622\textbf{} $|$ \textbf{.053} $|$ \textbf{.036} $|$ .046 	        & .655\textbf{} $|$ .180\textbf{} $|$ .170\textbf{} $|$ .196  \\\bottomrule
\end{tabular}
\caption{Performances of machine learning models on each data set. Higher F-1 score and accuracy (acc) are better, whereas for Equalized Opportunity ($\delta_{opp}$), Equalized Odd ($\delta_{odd}$), and Demographic Parity ($\delta_{demP}$), lower is better. Figures are in absolute values and truncated to three decimal places.}
\label{table:ml_performances}
\end{table}

\begin{figure}[h]
    \centering
     \begin{subfigure}{0.33\textwidth}
        {\includegraphics[width=\textwidth]{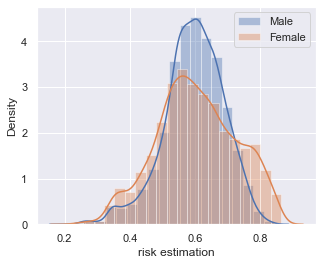}}
        \caption{Adult}
        \label{fig:risk_distribution_adult}
     \end{subfigure}
    \begin{subfigure}{0.33\textwidth}
       {\includegraphics[width=\textwidth]{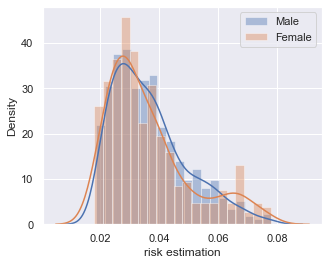}}
        \caption{Dropout}
        \label{fig:risk_distribution_dropout}
    \end{subfigure}
    \begin{subfigure}{0.33\textwidth}
       {\includegraphics[width=\textwidth]{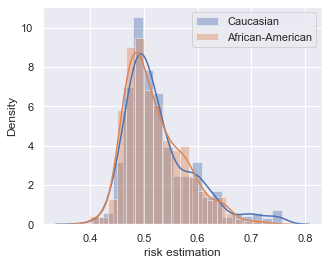}}
        \caption{Compas}
        \label{fig:risk_distribution_compas}
    \end{subfigure}
    \caption{Distribution of ex-ante risk score for Adults (a), Dropout (b) and Compas (c) dataset.}
    \label{fig:risk_distribution}
\end{figure}
 
Figure \ref{fig:risk_distribution} plots the distribution of FORESEE's estimates of ex-ante fairness risk. Across all datasets there exist instances with high risk scores, which indicates that misclassification rates would vary with changes in sensitive attribute. This is particularly striking for Compas, where whole of the population is estimated to have an ex-ante risk score larger than $\approx0.4$. High risk scores for Compas confirms findings in existing studies (e.g \cite{ProPublica2016}) that this dataset is not well suited for training automated decision systems without significantly harming a demographic group.

Using Figure \ref{fig:risk_distribution}, we define as high risk a sample with risk higher than $0.5$ for Adults and Compas, and higher than $0.06$ for Dropout. In Table \ref{table:ml_performances2}, we find that all three fairness metrics are higher for high risk individuals than low risk ones. This is evidence that our ex-ante risk score allows identifying individuals for which a given classifier's aggregate performances would significantly vary across demographic groups.  Notice that the differences are not significant for Compas as the other datasets, which is consistent with the distribution of our risk core estimates depicted in Figure \ref{fig:risk_distribution} and flags once again the issues of using this set for automated training \cite{bao2021s}.

\begin{table}[h]
\centering
\begin{tabular}{@{}llll@{}}
\toprule
  \textbf{Fairness }  & \textbf{Adult} & \textbf{Dropout} & \textbf{Compas}  \\
        \textbf{Metric}    & High \textbf{} $|$ Low & High \textbf{} $|$ Low & High \textbf{} $|$ Low\\\midrule
        \textbf{LR} \\
        $\delta_{opp}$    & 0.496 $|$ 0.003 &   0.500 $|$ 0.097   &   0.291 $|$ 0.164 \\
        $\delta_{odd}$    & 0.379 $|$ 0.031 &   0.250 $|$ 0.059   &   0.196 $|$ 0.194 \\
        $\delta_{demP}$   & 0.391 $|$ 0.068 &   0.146 $|$ 0.061   &   0.215 $|$ 0.206 \\ \midrule
        
        \textbf{RF} \\
        $\delta_{opp}$    & 0.246 $|$ 0.026 &   0.336 $|$ 0.173   &   0.306 $|$ 0.077 \\
        $\delta_{odd}$    & 0.247 $|$ 0.028 &   0.203 $|$ 0.096   &   0.203 $|$ 0.161 \\
        $\delta_{demP}$   & 0.372 $|$ 0.040 &   0.047 $|$ 0.063   &   0.222 $|$ 0.174 \\ \midrule
        
        \textbf{KNN} \\
        $\delta_{opp}$    & 0.231 $|$ 0.050 &   0.227 $|$ 0.043   &   0.320 $|$ 0.057 \\
        $\delta_{odd}$    & 0.167 $|$ 0.030 &   0.131 $|$ 0.029   &   0.206 $|$ 0.145 \\
        $\delta_{demP}$   & 0.225 $|$ 0.010 &   0.081 $|$ 0.042   &   0.220 $|$ 0.154 \\\midrule
        
        \textbf{SVM} \\
        $\delta_{opp}$    & 0.245 $|$ 0.009 &   0.145 $|$ 0.096   &   0.287 $|$ 0.091 \\
        $\delta_{odd}$    & 0.286 $|$ 0.020 &   0.126 $|$ 0.056   &   0.193 $|$ 0.159 \\
        $\delta_{demP}$   & 0.428 $|$ 0.043 &   0.031 $|$ 0.054   &   0.213 $|$ 0.170 \\ \bottomrule
\end{tabular}
\caption{Fairness metrics in absolute value for groups defined by level of ex-ante fairness risk score. The thresholds between Low and High risk groups are set to $0.5$ for Adult and Compas, and $0.06$ for Dropout. We truncate figures to three decimal places.}
\label{table:ml_performances2}
\end{table}

\subsection{RQ3: Profiles for High and Low Risk Instances}
\begin{figure}[h]
    \centering
     \begin{subfigure}{0.33\textwidth}
        {\includegraphics[width=\textwidth]{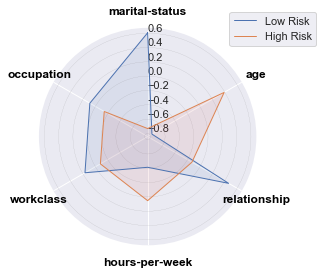}}
        \caption{Adult}
        \label{fig:adult_profiles}
     \end{subfigure}
    \begin{subfigure}{0.33\textwidth}
       {\includegraphics[width=\textwidth]{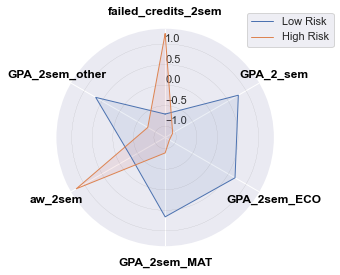}}
        \caption{Dropout}
        \label{fig:dropout_profiles}
    \end{subfigure}
    \begin{subfigure}{0.33\textwidth}
       {\includegraphics[width=\textwidth]{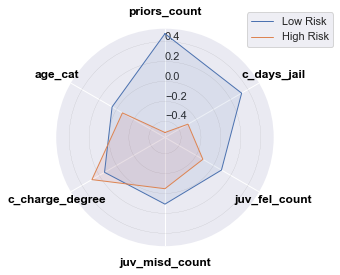}}
        \caption{Compas}
        \label{fig:compas_profiles}
    \end{subfigure}
    \caption{Characteristics of samples considered at high and low fairness risk by FORESEE.}
    \label{fig:bins_profiles}
\end{figure}

Figure \ref{fig:bins_profiles} shows the  differences between the High and Low Risk profiles in the datasets. 
The high risk group is composed of the top 20\% of samples with high risk scores drawn equally from both groups in $S$. The low risk group is similarly formed with the lowest 20\% risk scores. From figure \ref{fig:bins_profiles} we can characterizes high risks as in the following list (the low profile are the opposite to this description):

\begin{itemize}
    \item \emph{Adult}: aged individuals, high number of working hours per week, and married or divorced.
    \item \emph{Dropout}: students with low academic performances and higher numbers of academic warnings.
    \item \emph{Compas}: younger defendants, fewer criminal history, mostly felony as charge degree, and spending less days in jail.
\end{itemize}

Our risk score provides stakeholders and data owners information on the characteristics of individuals for which predictions would be affected by their sensitive attributes. A useful implication of the profiles in Figure \ref{fig:bins_profiles} is to use with caution any data mining algorithm that predicts the target variable if the individual' risk estimate is high.

\subsubsection{RQ4: Risk Score Guided- Mitigation Strategies.}
We test the pre- and post-processing strategies previously described.  Table \ref{table:ml_performances_mitigation} shows that a pre-processing strategy that removes  samples at high risk of unfair outcomes is effective at reducing aggregate inequality of odds and opportunities. This result is evidence that our ex-ante risk score is instrumental in identifying the data instances that push the classifier toward unfair outcomes. Moreover, our pre-processing strategy even improves the performance, which are consistent with previous studies that have shown results in favor of pre-processing approaches \cite{vasquez2022faired, khodadadian2021impact}. On the other hand, removing high risk individuals in Compas dataset results in insignificant improvement, which is expected given the found distribution of risk scores. Intuitively, although we remove instances with scores $>0.5$, the remaining still has high levels (mostly $>0.4$). Consequently, we find that post-processing approaches on the risk sub-sample might be better a approach, where post-processing subjected to demographic parity is the best one.

Table \ref{table:ml_performances_mitigation} also shows that post-processing tailored to high risk group is effective at reducing each fairness metric, although at a cost of performance. Therefore, 
developers can use the high risk group as a unit test against which they can automatically evaluate the fairness of their algorithm; then, post-process classification outcomes for that group. 

\begin{table}[h]
\centering
\begin{tabular}{@{}lllll@{}}
\toprule
  \textbf{Model}  & \textbf{Adult-RF} & \textbf{Dropout-LR} & \textbf{Compas-RF}  \\ 
  	& F-1 \textbf{ }$|$ $\delta_{opp}$ $|$ $\delta_{odd}$ $|$ $\delta_{demP}$ &  F-1 \textbf{ } $|$ $\delta_{opp}$ $|$ $\delta_{odd}$ $|$ $\delta_{demP}$ & acc \textbf{ } $|$ $\delta_{opp}$ $|$ $\delta_{odd}$ $|$ $\delta_{demP}$ \\ \midrule
  original  & .703 $|$ .200 $|$ .208 $|$ .331   & .655 $|$ .062 $|$ .038 $|$ .031   & .668 $|$ .178 $|$ .174 $|$ .204	\\ \midrule
  pre-processing (train\&test)	   & .794\textbf{} $|$ \textbf{.024} $|$ .033\textbf{} $|$ .056     & \textbf{.662} $|$ \textbf{.058} $|$ \textbf{.040} $|$ \textbf{.058}   & .578 \textbf{}$|$ .133\textbf{} $|$ .148\textbf{} $|$ .158\textbf{} 	\\
  pre-processing (test)	   & \textbf{.817} $|$ .026\textbf{} $|$ \textbf{.028} $|$ .040     & .655\textbf{} $|$ .097 \textbf{}$|$ .059 \textbf{}$|$ .061    & .625\textbf{} $|$ .077 \textbf{}$|$ .161\textbf{} $|$ .174 	\\
  post-processing (demP)   & .621 \textbf{}$|$ .066 \textbf{}$|$ .084\textbf{} $|$ \textbf{.035}    & .599 \textbf{}$|$ .169 \textbf{}$|$ .102 \textbf{}$|$ .074   & \textbf{.630} $|$ .064\textbf{} $|$ \textbf{.054} $|$ \textbf{.030}	\\
  post-processing (eqODD)  & .636 \textbf{}$|$ .076 \textbf{}$|$ .051\textbf{} $|$ .141     & .601 \textbf{}$|$ .169 \textbf{}$|$ .101 \textbf{}$|$ .072    & .563\textbf{} $|$ \textbf{.022} $|$ .065\textbf{} $|$ .073	\\
  post-processing (eqOPP)  & .671 \textbf{}$|$ .118 \textbf{}$|$ .151\textbf{} $|$ .279     & .603 \textbf{}$|$ .157 \textbf{}$|$ .092 \textbf{}$|$ .065    & .627\textbf{} $|$ .026\textbf{} $|$ .112\textbf{} $|$ .097	\\
  \bottomrule
\end{tabular}
\caption{Model performances and fairness metrics in absolute values after implementing pre and post-processing mitigation strategies. Larger F-1 score and lower fairness metrics show a better accuracy-fairness trade-off. Values are truncated to three decimal places.}
\label{table:ml_performances_mitigation}
\end{table}

\section{Conclusions}
\label{sec:conclusions}
Although the fair machine learning literature offers an abundance of  metrics to measure the potential harmful effect on individuals of a machine learning model, stakeholders would like to anticipate these fairness pitfalls before investing in model development. We propose an ex-ante risk score to measure whether future performances of any classifier using the data will vary with sensitive attributes. Moreover, we propose a bagging approach (\emph{FORESEE}) to estimate the risk score from a finite sample. Our experiments demonstrate that our risk score, although measured before model development, anticipates correctly whether a model using the data will under-serve some demographic groups. Moreover, our risk score offers useful guidance for stakeholders to decide whether a data is appropriate for training automated decision systems; for developers to mitigate potentially discriminatory outcomes; and for end-users to decide when to trust the outcome of the algorithm. However, our approach is still computational expensive. Also, although a pessimistic scenario assumption may over-estimate the misclassification rate variation, our results show this seems to be sufficient.

\section*{Acknowledgments}
This study was supported by the National Agency for Research and Development (ANID - Agencia Nacional de Investigación y Desarrollo/Subdirección de Capital Humano), "Becas Chile" Doctoral Fellowship 2020 program; Grant No. 72210492 to Jonathan Patricio Vasquez Verdugo.

\bibliographystyle{unsrt}  
\bibliography{references}

\begin{thebibliography}{10}

\bibitem{green2019disparate}
Ben Green and Yiling Chen.
\newblock Disparate interactions: An algorithm-in-the-loop analysis of fairness
  in risk assessments.
\newblock In {\em Proceedings of the conference on fairness, accountability,
  and transparency}, pages 90--99, 2019.

\bibitem{ProPublica2016}
ProPublica.
\newblock How we analyzed the compas recidivism algorithm.
\newblock {\em ProPublica}, 2016.

\bibitem{fu2021crowds}
Runshan Fu, Yan Huang, and Param~Vir Singh.
\newblock Crowds, lending, machine, and bias.
\newblock {\em Information Systems Research}, 32(1):72--92, 2021.

\bibitem{wen2019fairness}
Min Wen, Osbert Bastani, and Ufuk Topcu.
\newblock Fairness with dynamics.
\newblock {\em arXiv preprint arXiv:1901.08568}, 2019.

\bibitem{chen2021algorithm}
Richard~J Chen, Tiffany~Y Chen, Jana Lipkova, Judy~J Wang, Drew~FK Williamson,
  Ming~Y Lu, Sharifa Sahai, and Faisal Mahmood.
\newblock Algorithm fairness in ai for medicine and healthcare.
\newblock {\em arXiv preprint arXiv:2110.00603}, 2021.

\bibitem{seyyed2020chexclusion}
Laleh Seyyed-Kalantari, Guanxiong Liu, Matthew McDermott, Irene~Y Chen, and
  Marzyeh Ghassemi.
\newblock Chexclusion: Fairness gaps in deep chest x-ray classifiers.
\newblock In {\em BIOCOMPUTING 2021: Proceedings of the Pacific Symposium},
  pages 232--243. World Scientific, 2020.

\bibitem{kizilcec2020algorithmic}
Ren{\'e}~F Kizilcec and Hansol Lee.
\newblock Algorithmic fairness in education.
\newblock {\em arXiv preprint arXiv:2007.05443}, 2020.

\bibitem{buolamwini2018gender}
Joy Buolamwini and Timnit Gebru.
\newblock Gender shades: Intersectional accuracy disparities in commercial
  gender classification.
\newblock In {\em Conference on fairness, accountability and transparency},
  pages 77--91. PMLR, 2018.

\bibitem{aif360-oct-2018}
Rachel K.~E. Bellamy, Kuntal Dey, Michael Hind, Samuel~C. Hoffman, Stephanie
  Houde, Kalapriya Kannan, Pranay Lohia, Jacquelyn Martino, Sameep Mehta,
  Aleksandra Mojsilovic, Seema Nagar, Karthikeyan~Natesan Ramamurthy, John
  Richards, Diptikalyan Saha, Prasanna Sattigeri, Moninder Singh, Kush~R.
  Varshney, and Yunfeng Zhang.
\newblock {AI Fairness} 360: An extensible toolkit for detecting,
  understanding, and mitigating unwanted algorithmic bias, October 2018.

\bibitem{bird2020fairlearn}
Sarah Bird, Miro Dud{\'i}k, Richard Edgar, Brandon Horn, Roman Lutz, Vanessa
  Milan, Mehrnoosh Sameki, Hanna Wallach, and Kathleen Walker.
\newblock Fairlearn: A toolkit for assessing and improving fairness in {AI}.
\newblock Technical Report MSR-TR-2020-32, Microsoft, May 2020.

\bibitem{gitiaux2021fair}
Xavier Gitiaux and Huzefa Rangwala.
\newblock Fair representations by compression.
\newblock In {\em Proceedings of the AAAI Conference on Artificial
  Intelligence}, volume~35, pages 11506--11515, 2021.

\bibitem{alla2021mlops}
Sridhar Alla and Suman~Kalyan Adari.
\newblock What is mlops?
\newblock In {\em Beginning MLOps with MLFlow}, pages 79--124. Springer, 2021.

\bibitem{pessach2020algorithmic}
Dana Pessach and Erez Shmueli.
\newblock Algorithmic fairness.
\newblock {\em arXiv preprint arXiv:2001.09784}, 2020.

\bibitem{akpinar2022sandbox}
Nil-Jana Akpinar, Manish Nagireddy, Logan Stapleton, Hao-Fei Cheng, Haiyi Zhu,
  Steven Wu, and Hoda Heidari.
\newblock A sandbox tool to bias (stress)-test fairness algorithms.
\newblock {\em arXiv preprint arXiv:2204.10233}, 2022.

\bibitem{chen2018my}
Irene Chen, Fredrik~D Johansson, and David Sontag.
\newblock Why is my classifier discriminatory?
\newblock {\em Advances in Neural Information Processing Systems}, 31, 2018.

\bibitem{ensign2018runaway}
Danielle Ensign, Sorelle~A Friedler, Scott Neville, Carlos Scheidegger, and
  Suresh Venkatasubramanian.
\newblock Runaway feedback loops in predictive policing.
\newblock In {\em Conference on Fairness, Accountability and Transparency},
  pages 160--171. PMLR, 2018.

\bibitem{gebru2021datasheets}
Timnit Gebru, Jamie Morgenstern, Briana Vecchione, Jennifer~Wortman Vaughan,
  Hanna Wallach, Hal~Daum{\'e} Iii, and Kate Crawford.
\newblock Datasheets for datasets.
\newblock {\em Communications of the ACM}, 64(12):86--92, 2021.

\bibitem{adkins2022prescriptive}
David Adkins, Bilal Alsallakh, Adeel Cheema, Narine Kokhlikyan, Emily
  McReynolds, Pushkar Mishra, Chavez Procope, Jeremy Sawruk, Erin Wang, and
  Polina Zvyagina.
\newblock Prescriptive and descriptive approaches to machine-learning
  transparency.
\newblock In {\em CHI Conference on Human Factors in Computing Systems Extended
  Abstracts}, pages 1--9, 2022.

\bibitem{mitchell2019model}
Margaret Mitchell, Simone Wu, Andrew Zaldivar, Parker Barnes, Lucy Vasserman,
  Ben Hutchinson, Elena Spitzer, Inioluwa~Deborah Raji, and Timnit Gebru.
\newblock Model cards for model reporting.
\newblock In {\em Proceedings of the conference on fairness, accountability,
  and transparency}, pages 220--229, 2019.

\bibitem{chipman2010bart}
Hugh~A Chipman, Edward~I George, and Robert~E McCulloch.
\newblock Bart: Bayesian additive regression trees.
\newblock {\em The Annals of Applied Statistics}, 4(1):266--298, 2010.

\bibitem{feldman2015certifying}
Michael Feldman, Sorelle~A Friedler, John Moeller, Carlos Scheidegger, and
  Suresh Venkatasubramanian.
\newblock Certifying and removing disparate impact.
\newblock In {\em proceedings of the 21th ACM SIGKDD international conference
  on knowledge discovery and data mining}, pages 259--268, 2015.

\bibitem{dwork2011fairness}
Cynthia Dwork, Moritz Hardt, Toniann Pitassi, Omer Reingold, and Richard~S
  Zemel.
\newblock Fairness through awareness. corr abs/1104.3913 (2011).
\newblock {\em arXiv preprint arXiv:1104.3913}, 2011.

\bibitem{ilvento2019metric}
Christina Ilvento.
\newblock Metric learning for individual fairness.
\newblock {\em arXiv preprint arXiv:1906.00250}, 2019.

\bibitem{gitiaux2019multi}
Xavier Gitiaux and Huzefa Rangwala.
\newblock Multi-differential fairness auditor for black box classifiers.
\newblock {\em arXiv preprint arXiv:1903.07609}, 2019.

\bibitem{kearns2018preventing}
Michael Kearns, Seth Neel, Aaron Roth, and Zhiwei~Steven Wu.
\newblock Preventing fairness gerrymandering: Auditing and learning for
  subgroup fairness.
\newblock In {\em International Conference on Machine Learning}, pages
  2564--2572. PMLR, 2018.

\bibitem{kim2019multiaccuracy}
Michael~P Kim, Amirata Ghorbani, and James Zou.
\newblock Multiaccuracy: Black-box post-processing for fairness in
  classification.
\newblock In {\em Proceedings of the 2019 AAAI/ACM Conference on AI, Ethics,
  and Society}, pages 247--254, 2019.

\bibitem{kearns2019empirical}
Michael Kearns, Seth Neel, Aaron Roth, and Zhiwei~Steven Wu.
\newblock An empirical study of rich subgroup fairness for machine learning.
\newblock In {\em Proceedings of the Conference on Fairness, Accountability,
  and Transparency}, pages 100--109, 2019.

\bibitem{hill2011bayesian}
Jennifer~L Hill.
\newblock Bayesian nonparametric modeling for causal inference.
\newblock {\em Journal of Computational and Graphical Statistics},
  20(1):217--240, 2011.

\bibitem{asuncion2007uci}
Arthur Asuncion and David Newman.
\newblock Uci machine learning repository, 2007.

\bibitem{hardt2016equality}
Moritz Hardt, Eric Price, and Nati Srebro.
\newblock Equality of opportunity in supervised learning.
\newblock {\em Advances in neural information processing systems}, 29, 2016.

\bibitem{bao2021s}
Michelle Bao, Angela Zhou, Samantha Zottola, Brian Brubach, Sarah Desmarais,
  Aaron Horowitz, Kristian Lum, and Suresh Venkatasubramanian.
\newblock It's compaslicated: The messy relationship between rai datasets and
  algorithmic fairness benchmarks.
\newblock {\em arXiv preprint arXiv:2106.05498}, 2021.

\bibitem{vasquez2022faired}
Jonathan Vasquez~Verdugo, Xavier Gitiaux, Cesar Ortega, and Huzefa Rangwala.
\newblock Faired: A systematic fairness analysis approach applied in a higher
  educational context.
\newblock In {\em LAK22: 12th International Learning Analytics and Knowledge
  Conference}, pages 271--281, 2022.

\bibitem{khodadadian2021impact}
Sajad Khodadadian, AmirEmad Ghassami, and Negar Kiyavash.
\newblock Impact of data processing on fairness in supervised learning.
\newblock In {\em 2021 IEEE International Symposium on Information Theory
  (ISIT)}, pages 2643--2648. IEEE, 2021.

\end{thebibliography}

\end{document}